\documentclass[sigplan,10pt,screen]{acmart}
\renewcommand\footnotetextcopyrightpermission[1]{}

\usepackage{placeins}
\usepackage{xspace}

\newcommand{\sys}{OneLA\xspace}
\AtBeginDocument{%
  }

\setcopyright{none}
\begin{document}

\title{\sys: Scaling Linear-Attention Decoding to Large Beams in Generative Recommendation}

\author{Xiangrui Yang}
\authornote{Xiangrui Yang and Cheng Peng contributed equally to this work.}
\affiliation{%
  \institution{The University of Hong Kong}
  \country{China}
}
\email{xiangruiyang@connect.hku.hk}

\author{Cheng Peng}
\authornotemark[1]
\affiliation{%
  \institution{Kuaishou Technology}
  \country{China}
}
\email{pengcheng09@kuaishou.com}

\author{Yunfeng Zhao}
\affiliation{%
  \institution{Kuaishou Technology}
  \country{China}
}
\email{zhaoyunfeng@kuaishou.com}

\author{Liang Zeng}
\affiliation{%
  \institution{Kuaishou Technology}
  \country{China}
}
\email{zengliang03@kuaishou.com}

\author{Ao Hu}
\affiliation{%
  \institution{Kuaishou Technology}
  \country{China}
}
\email{huao@kuaishou.com}

\author{Jiawei Yang}
\authornote{This work was completed at Kuaishou.}
\affiliation{%
  \institution{Peking University}
  \country{China}
}
\email{yangjiawei26@stu.pku.edu.cn}

\author{Shengzhe Wang}
\affiliation{%
  \institution{Kuaishou Technology}
  \country{China}
}
\email{wangshengzhe@kuaishou.com}

\author{Jingshan Lv}
\affiliation{%
  \institution{Kuaishou Technology}
  \country{China}
}
\email{lvjingshan@kuaishou.com}

\author{Xiao Liang}
\affiliation{%
  \institution{Kuaishou Technology}
  \country{China}
}
\email{liangxiao03@kuaishou.com}

\author{Chen Yang}
\affiliation{%
  \institution{Kuaishou Technology}
  \country{China}
}
\email{yangchen28@kuaishou.com}

\author{Jiaqiang Liu}
\authornote{Jiaqiang Liu and Yiming Qiu are the corresponding authors.}
\affiliation{%
  \institution{Kuaishou Technology}
  \country{China}
}
\email{liujiaqiang@kuaishou.com}

\author{Yiming Qiu}
\authornotemark[3]
\affiliation{%
  \institution{The University of Hong Kong}
  \country{China}
}
\email{yimingq@hku.hk}

\begin{abstract}

Generative recommendation (GR) relies on large-beam decoding to generate hundreds of candidate items, creating a new scaling challenge for recurrent linear attention.
Existing linear attention serving systems either materialize a full recurrent state for every beam or repeatedly replay shared history, incurring substantial memory and traffic overhead.

To address this, we present \sys, a linear-attention decoding framework that exploits the shared prompt and short divergent suffixes of GR workloads.
Specifically, \sys represents all beam states using a single shared prompt-derived state and compact, append-only records of their divergent transitions.
Using this representation, \sys computes only the state information required at each decoding step, without reconstructing a full recurrent state for every beam.
Furthermore, \sys uses a lightweight ancestry index to track the transition records that make up each beam's history, allowing beams to be updated without moving or copying existing records.
A fused GPU kernel further reuses the shared state across beams.
Our analysis shows that \sys achieves 1.54--2.46$\times$ end-to-end decode speedups while substantially reducing recurrent-state memory use and data movement.

\end{abstract}


\maketitle
\fancyhf{}\fancyfoot[C]{\footnotesize\thepage}\renewcommand{\headrulewidth}{0pt}

\section{Introduction}
\label{sec:introduction}

Recommendation systems are central to modern online services, and generative recommendation (GR) has emerged as a practical paradigm for large-scale applications~\cite{onerec2026onereason, zhou2025onerec, he2026plum, zhai2024hstu, chen2026onesearch}.
The widely adopted approach represents each item as a short sequence of discrete semantic IDs (SIDs), thereby converting recommendation into sequence generation while retaining semantic and collaborative structure~\cite{rajput2023tiger,zheng2024lcrec,wang2024letter,wang2024eager,hou2025actionpiece,adaptive_semantic_quantization}.
Recent industrial systems such as OneRec and OneReason illustrate both the adoption and continued extension of this formulation~\cite{zhou2025onerec,onerec2026onereason}.
Unlike conventional LLM serving, GR must produce hundreds of candidate items rather than a single sequence and therefore relies on large-beam decoding, where candidate SID sequences are expanded and globally selected at every step~\cite{sun2025xgr}.
The resulting workload combines a long shared prompt, short generated suffixes, and a large, dynamically evolving beam.

Processing such long user histories~\cite{douyin_long_sequence} with conventional attention incurs quadratic computation in sequence length, which motivates the use of linear attention in GR~\cite{katharopoulos2020transformers}.
These modules compress token history into a fixed-size recurrent state, substantially reducing the dependence of computation and memory on context length.
We use Gated DeltaNet (GDN)~\cite{yang2025gated}, widely adopted in frontier LLMs~\cite{qwen3next}, as our representative instance.
However, large-beam GR turns this fixed-size state into a new scaling problem: an independently evolving state for every live beam makes storage and memory traffic grow sharply with beam width.

General-purpose LLM serving systems optimize KV-cache management and attention execution for conventional decoding, where each sequence owns an independent state~\cite{kwon2023vllm,zheng2023sglang,ye2025flashinfer,trtllm}.
Their GDN execution paths inherit this assumption: a beam either persists its own complete recurrent state or reconstructs one from its own history, and in both cases the state is bound to the logical beam that owns it~\cite{kwon2023vllm,ye2025flashinfer,trtllm}.
This binding is particularly costly in GR, where global beam selection reorders and duplicates beams at every decoding step, forcing states to be copied, reconstructed, or recomputed as the beam tree evolves~\cite{sun2025xgr}.

We observe that full per-beam recurrent state materialization is fundamentally redundant.
In large-beam GR, all candidate beams within a request originate from an identical prompt-derived recurrent state and diverge only through a brief sequence of decode-time state transitions.
Each transition can be represented compactly, allowing a beam state to be expressed as one shared post-prefill state together with a short sequence of branch-specific transition records.
This trades computation for memory, with replay depth bounded by the fixed-length SID sequence.
Moreover, beam selection changes only which historical records a surviving beam refers to; it does not modify the records themselves. These records can therefore remain immutable and append-only, while lightweight ancestry metadata tracks the dynamically evolving beam tree.

Based on these observations, we present \sys, an efficient linear-attention decoding framework for large-beam GR.
\sys stores the post-prefill recurrent state once per request and represents beam-specific evolution with compact GDN Transition Records (GTRs).
Rather than using these records to reconstruct each beam's recurrent state, \sys replays them onto the two state projections that each decoding step actually uses.
Replay therefore costs vector rather than matrix processing, and no per-beam recurrent matrix is ever materialized or written back.
A lightweight ancestry index decouples dynamic beam evolution from physical record placement, while a fused shared-context GPU kernel reuses the shared state across beams and performs replay entirely on chip.
We also implement the \sys prototype and evaluate it on an industrial GR workload.
The results show that \sys achieves 1.54--2.46$\times$ end-to-end decode speedups over existing GDN execution paths.
In summary, this paper makes the following contributions:
\begin{itemize}

\item We identify beam width as a new scaling dimension for recurrent linear attention in GR, exposing the inefficiency of per-beam state management.
\item We design \sys with a shared context state, compact append-only GDN Transition Records (GTRs), and lightweight beam ancestry, and develop a fused shared-context GPU kernel that performs on-chip projection replay with cross-beam context-state reuse.
\item We implement and evaluate \sys, which achieves 1.54--2.46$\times$ end-to-end decode speedups over existing GDN execution paths.
\end{itemize}

\section{Background and Motivation}
\label{sec:background}

\subsection{Generative Recommendation}

Generative recommendation (GR) reformulates candidate generation as autoregressive sequence generation~\cite{rajput2023tiger,zhou2025onerec,holorec}.
Recent systems, such as OneRec, OneReason, PLUM, and S-GRec, have established SID-based generation as a common GR formulation~\cite{10.1145/3770855.3818473,zhou2025onerec,he2026plum,onerec2026onereason}: each item is represented by a short sequence of discrete semantic IDs (SIDs), which the model generates.
Formally, an item $i$ is represented as $\mathbf{s}_i=(s_i^1,\ldots,s_i^T)$.
Given an input prompt $\mathbf{x}$, the model generates its SID autoregressively: $\prod_{t=1}^{T} P(s_i^t\mid\mathbf{x},s_i^{<t})$.

After the generation process, the complete SID sequence is mapped back to an item~\cite{rajput2023tiger}. 
Unlike conventional text generation, GR must produce hundreds of candidate items rather than a single best sequence~\cite{sun2025xgr}. It therefore employs large-beam decoding to explore SID sequences.

\begin{figure}[t]
  \centering
  \includegraphics[width=0.6\linewidth]{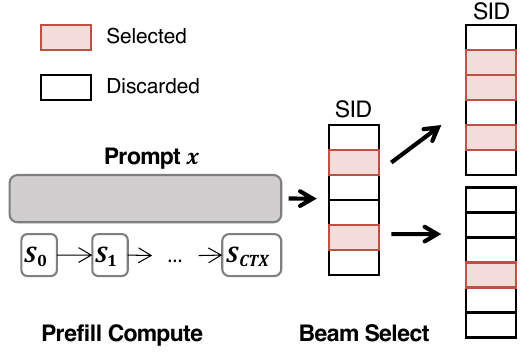}
  \vspace{-5mm}
  \caption{Decoding workflow for GR.}
  \Description{A prompt is processed during prefill to produce a shared context state. Beam selection then expands short semantic-ID sequences, retaining multiple candidates that can share a parent prefix.}
  \label{fig:gr-workflow}
  \vspace{-7mm}
\end{figure}

At each decoding step, all active beams are expanded over the SID codebook, and the global top-$k$ candidates are retained~\cite{sun2025xgr}.
Multiple surviving children may originate from the same parent, while other parents may disappear.
Consequently, the parent mapping $\pi_t(b)$ of beam $b$ changes across steps, forming a dynamic beam tree.
The resulting workload combines a long shared prompt, a short generated SID sequence, and a large dynamically branching beam~\cite{sun2025xgr}.
That is to say, the prompt length and beam width are much greater than the SID sequence length, i.e., $L_{prompt} \gg T$ and $W_t \gg T$~\cite{sun2025xgr}.
Figure~\ref{fig:gr-workflow} summarizes this workflow: prefill computes the shared prompt, after which beam selection expands and retains candidate SID sequences.

\subsection{Gated Delta Network}
The long prompts of this workload are costly for conventional attention, whose computation is quadratic in sequence length and whose KV cache grows linearly with it~\cite{vaswani2017attention,katharopoulos2020transformers}.
Linear attention, such as GDN, instead uses a fixed-size recurrent state: each head compresses its preceding history into $S_t\in\mathbb{R}^{d_v\times d_k}$, updated once per token, where $d_v$ and $d_k$ denote the value and key dimensions.
Specifically, GDN processes token $t$ from state $S_{t-1}$ as follows~\cite{yang2025gated}:

\begin{equation}
\begin{aligned}
S_t
&= \alpha_t S_{t-1}
+ \beta_t
\left(
v_t - \alpha_t S_{t-1}{k}_t
\right)
{k}_t^{T},\\
o_t
&= S_t q_t,
\end{aligned}
\label{eq:next_state}
\end{equation}
where $\alpha_t$ and $\beta_t$ control state decay and correction, respectively, and $q_t$ and $k_t$ denote the query and key.
We refer to one such update, $S_{t-1}\!\to\!S_t$, as a \emph{GDN transition step}. 
The state $S_t$ summarizes the history up to token $t$, and the output $o_t$ reads it out along the query $q_t$.

\subsection{GDN Decoding for Generative Recommendation}

Linear attention like GDN is widely adopted in frontier LLMs~\cite{qwen3next} and works well for GR models that need to absorb long user contexts.
However, existing LLM serving frameworks handle GDN-based large-beam GR inefficiently.
The difficulty arises when the compact recurrent representation of GDN is combined with the large and dynamically branching beam in GR.

\begin{figure}[t]
  \centering
  \includegraphics[width=\linewidth]{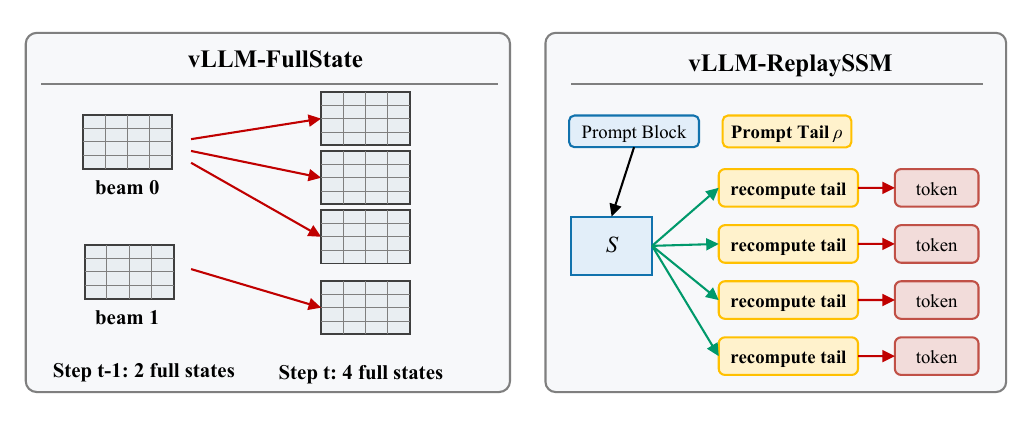}
  \vspace{-6mm}
  \caption{State management in an existing framework.}
  \Description{The FullState path expands two per-beam recurrent-state
  matrices at step t-minus-one into four matrices at step t.  The ReplaySSM
  path shares a cached prompt-block state, but independently recomputes the
  prompt tail before processing each beam's next token.}
  \vspace{-8mm}
  \label{fig:vllm-state-management}
\end{figure}

Among existing serving frameworks, only vLLM~\cite{kwon2023vllm} provides native end-to-end support for GDN-based beam search.
As shown in Figure~\ref{fig:vllm-state-management}, the FullState path in vLLM maintains a complete recurrent state for every live beam, resulting in many huge full-state matrices.
The ReplaySSM path in vLLM reduces state persistence through prefix replay but does not preserve replay history across the independent requests created by the native beam-search path.
Specifically, ReplaySSM does not explicitly distinguish the long shared prompt from the short decode phase: it materializes a recurrent-state checkpoint only at completed prompt-block boundaries, potentially leaving a large prompt tail to be replayed during decoding. 
Moreover, ReplaySSM treats each beam as an independent request, requiring every beam to separately load the cached state checkpoint and replay its subsequent tokens.
Other frameworks, including SGLang, FlashInfer, and TensorRT-LLM, provide operator-level support for GDN modules but do not offer end-to-end GDN serving with beam search.
Overall, existing systems either materialize recurrent state independently for each beam or lack native end-to-end support for this workload.

Existing state-management approaches become inefficient under this workload for two main reasons.
First, although GDN eliminates sequence-length-dependent KV storage~\cite{katharopoulos2020transformers,yang2025gated}, the large-beam search decoding process of GR introduces a new scaling dimension~\cite{sun2025xgr}.
Maintaining an independent state for $w$ live beams requires $\Theta(wd_vd_k)$ persistent state capacity.
The per-beam dimension also amplifies recurrent state reads and writebacks during every decoding step.
Second, dynamic beam evolution becomes expensive when materialized states are bound directly to logical beams.
At every decoding step, global beam selection may reorder or duplicate beams and change the beam width~\cite{sun2025xgr}.
Therefore, when materialized states are bound directly to individual beams, these dynamic parent--child reassignments require state copying, reconstruction, or recomputation.

These limitations expose a mismatch between logical beam evolution and existing materialized-state management.
Efficient large-beam GDN decoding should therefore avoid materializing a complete recurrent state for every beam and decouple logical beam evolution from materialized state placement.
We next identify the key observations that enable such a representation.

\subsection{Key Insight}

The above limitations stem from the assumption that each logical beam needs to materialize a recurrent state or be treated as an independent request.
However, we derive two key insights that enable more efficient state management.

First, the recurrent states of different beams share the same prompt-derived origin and diverge only through a short sequence of GDN transition steps.
Moreover, each GDN transition step can be represented by a compact tuple whose size is substantially smaller than the full recurrent state.
More importantly, each such step consumes the recurrent state only through two projections, so the state itself never has to be reconstructed.

Second, the beam selection process only keeps references to prefix beams rather than changing the original prefix data.
The selection process may reorder or duplicate parents but does not modify previously generated recurrent updates. 
Therefore, these records can remain append-only, and lightweight ancestry information can specify which records belong to each selected beam, avoiding state or  record relocation as beam selection proceeds.
This decouples logical beam selection from materialized state placement and enables efficient sharing across dynamically branching beams.

These insights motivate a recurrent-state management approach that separates the shared prompt-derived state, branch-specific compact records, and logical beam ancestry, as we show in the following sections.

\section{Design}
\label{sec:design}

We present \sys, a GDN attention decode framework for industrial GR.
Following the above key observation that all beams within a request share an identical prompt-derived state and differ only in their short decode suffixes, \sys leverages the insights that GDN transitions are fully captured by compact tuples and that dynamic beam evolution can be tracked via lightweight ancestry over immutable prefix records.
\sys therefore decomposes state management into a single shared post-prefill context alongside append-only branch records, evaluating subsequent tokens via projection replay without materializing per-beam recurrent matrices.

\subsection{Design Overview}
\label{sec:design-overview}

For the GR decoding process, we use $t$ to denote the current decoding step, $W_t$ the number of active beams at step $t$, and $b \in \{0, \dots, W_t - 1\}$ a beam at that step. 
For $t > 0$, $\pi_t(b)$ denotes the parent of beam $b$ selected from the preceding step.
We denote by $S_{\mathrm{ctx}}$ the recurrent state produced after prompt prefill, which is shared by all beams within the same request.
For exposition, $S_{t, b}$ denotes the conceptual recurrent state of beam $b$ after decode step $t$, which is not materialized in \sys.
Unless noted, we describe one request and one GDN head and omit request and head indices for clarity.

\begin{figure}[t]
  \centering
  \includegraphics[width=\linewidth]{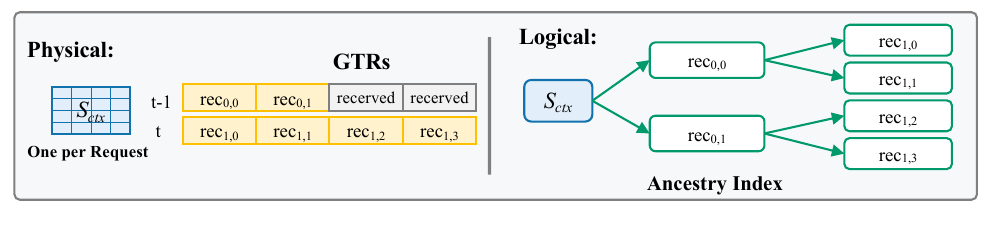}
  \vspace{-8mm}
  \caption{\sys Overview}
  \vspace{-6mm}
  \label{fig:design-overview}
\end{figure}

As shown in Figure~\ref{fig:design-overview}, \sys represents each beam as the shared post-prefill state $S_{\mathrm{ctx}}$ together with a short sequence of branch-specific state transitions, instead of materializing an independent recurrent state $S_{t,b}$ for every beam.
Based on this representation, \sys consists of two key components.
First, it retains $S_{\mathrm{ctx}}$ and reconstructs the state-dependent computation of each beam by replaying only its short sequence of branch-specific transitions.
Second, \sys decouples logical beam evolution from physical state storage through lightweight beam ancestry.
Together, these components allow \sys to share the prompt-derived state across beams while maintaining only lightweight branch-specific information as the beam tree evolves dynamically.

\subsection{\sys State Representation}
\label{sec:design-replay}

To avoid materialization of a full recurrent state for every beam, \sys represents each beam’s recurrent evolution using a shared post-prefill state together with a short sequence of compact GDN Transition Records (GTRs).
During decoding, \sys replays these records directly on the state projections, reproducing the original GDN computation.
For clarity, we omit the beam index $b$ throughout this subsection.

\noindent\textbf{Compact update records.}
In \sys, at decoding step $t$, the GDN recurrent update can be written as
\begin{equation}
\begin{aligned}
S_t
&=
\alpha_t S_{t-1}
+
\beta_t
\left(
v_t
-
\alpha_t S_{t-1} k_t
\right) k_t^{\top}
\\
&=
\alpha_t S_{t-1}
+
\delta_t k_t^{\top},
\end{aligned}
\label{eq:state_update}
\end{equation}
where $S_{t-1}$ is the state of the selected parent beam, with the shared post-prefill state $S_{\mathrm{ctx}}$ serving as the initial state.
We refer to $r_t = (\alpha_t, \delta_t, k_t)$ as a \textbf{GDN Transition Record} (GTR), in which $\delta_t = \beta_t\left(v_t-\alpha_t S_{t-1} k_t\right)$.
Given the parent state, a GTR completely determines the transition to the child state.

In our design, \sys stores $S_{\mathrm{ctx}}$ only once and records each GDN transition step by appending one GTR for each generated beam candidate, instead of persistently materializing a full $d_v\times d_k$ recurrent matrix ($O(d_vd_k)$) for every active beam as conventional FullState designs do.
Because a GTR consists only of a scalar decay factor, a $d_v$-dimensional update vector, and a $d_k$-dimensional key vector, each transition requires merely $O(d_v+d_k)$ storage.
We store $\delta_t$ rather than the underlying $(\beta_t, v_t)$ because replay never needs the state matrix itself, only its projections onto the current query and key, to which a record contributes $\delta_t$ scaled by a single dot product.
Keeping $(\beta_t, v_t)$ instead leaves the contribution dependent on the parent state $S_{t-1}$, which would have to be reconstructed before the record could be used, defeating the purpose of the record.

\begin{figure}[t]
  \centering
  \includegraphics[width=0.9\linewidth]{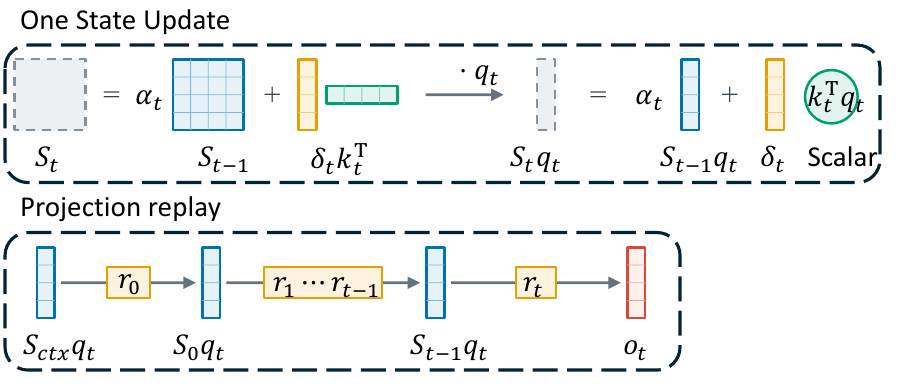}
  \vspace{-4mm}
  \caption{Projection replay in vector form.}
  \Description{A single GDN state update, drawn as a decayed matrix plus a
  rank-one outer product, becomes a vector plus a scalar-scaled vector once it is
  right-multiplied by the query; applying the same identity along a beam
  ancestry replays the stored transition records into the current output.}
  \vspace{-6mm}
  \label{fig:projection-replay}
\end{figure}

\noindent\textbf{Projection replay}
Although representing beams as GTR sequences eliminates persistent matrix storage, computing the subsequent decode step still requires the historical state context.
Crucially, a GDN layer interacts with the prior state $S_{t-1}$ solely through two vector contractions: the key projection $S_{t-1}k_t$ (for the update $\delta_t$) and the query projection $S_{t-1}q_t$ (for the output $o_t$).
Rather than reconstructing the intermediate $d_v \times d_k$ matrix on the fly, \sys performs \emph{projection replay} to evaluate these two projections directly across the ancestral chain of GTRs, as illustrated in Figure~\ref{fig:projection-replay}.

At decode step $t$, consider the sequence of GTRs $r_0,\ldots,r_{t-1}$ along the ancestry of the current beam.
For the current query $q_t$ and key $k_t$, we define a \emph{query-projection accumulator} $u_q^{(j)} \equiv S_j q_t$ and a \emph{key-projection accumulator} $u_k^{(j)} \equiv S_j k_t$, where $j$ denotes the replay depth along the beam ancestry.
Specifically, right-multiplying the state transition in Equation~\ref{eq:state_update} by $q_t$ or $k_t$ gives the following recurrences.
Starting from the shared context,
\begin{equation}
\begin{aligned}
u_q^{(-1)}
&=
S_{\mathrm{ctx}}q_t,
&
u_q^{(j)}
&=
\alpha_j u_q^{(j-1)}
+
\delta_j \left(k_j^{\top}q_t\right),
\\
u_k^{(-1)}
&=
S_{\mathrm{ctx}}k_t,
&
u_k^{(j)}
&=
\alpha_j u_k^{(j-1)}
+
\delta_j
\left(k_j^{\top}k_t\right),
\end{aligned}
 0 \le j < t.
\label{eq:projection_replay}
\end{equation}

After replaying through $j=t-1$, the accumulators directly provide the two projections required by the current GDN computation.
The current GDN update and output can therefore be evaluated as
\begin{equation}
\begin{aligned}
\delta_t
&=
\beta_t
\left(
v_t
-
\alpha_t u_k^{(t-1)}
\right),
\\
o_t
&=
\alpha_t u_q^{(t-1)}
+
\delta_t
\left(k_t^{\top}q_t\right).
\end{aligned}
\label{eq:replayed_update_output}
\end{equation}

Thus, projection replay requires only two $d_v$-dimensional accumulators per beam and value head, without materializing any intermediate $d_v\times d_k$ recurrent state.

\subsection{Dynamic Beam Ancestry Index}
\label{sec:design-ancestry}

\sys's state representation uses a shared context state $S_{\mathrm{ctx}}$ together with a short sequence of ancestral GTRs to replay a candidate beam.
Under dynamic beam search in the real world, however, global beam selection may prune, reorder, or fan out beams, so a logical beam can hardly be bound to a fixed set of physical records.
As pointed out previously, these operations do not modify the previously generated transition records and only update which records are referenced by each surviving beam.
\sys therefore keeps GTRs append-only in their original physical slots and tracks dynamic beam evolution separately through a lightweight \emph{ancestry index}.

Specifically, for beam $b$ at decoding step $t$, we define $h_t(b,j)$ as the physical slot for the GTR at decode step $j$, $j<t$, along the ancestry of $b$.
Let $\pi_t(b)$ denote the parent of beam $b$.
The ancestry index is updated as follows:
\begin{equation}
 h_t(b,j)=
 \begin{cases}
 h_{t-1}(\pi_t(b),j), & 0\le j<t-1,\\
 \pi_t(b),            & j=t-1.
 \end{cases}
 \label{eq:ancestry}
\end{equation}

Each child reuses the ancestry indices of its selected parent and records the parent’s physical slot for the immediately preceding decoding step.
Newly generated GTRs are written once to the physical slots of the current decoding step and are never relocated afterward. 
The logical beam tree is therefore represented entirely by the ancestry indices, while the underlying GTRs remain immutable.

This indirection decouples logical beam evolution from physical record placement.
Specifically, beam pruning discards references, reordering changes their order, and fan-out duplicates only ancestry indices rather than copying the corresponding GTR chains. 
Consequently, dynamic beam selection requires no movement or duplication of historical GTRs.
The projection replay can directly retrieve the record at depth $j$ as $r_{j,h_t(b,j)}$, obtaining the exact GTR sequence associated with each current beam.

\subsection{Fused Shared-Context Attention Kernel}
\label{sec:design-kernel}

The compact state representation removes per-beam full-state persistence, but its memory benefits must also be preserved during GPU execution.
In particular, $S_{\mathrm{ctx}}$ is shared across beams and should not be repeatedly fetched from HBM, while the transient projections used during replay should not be materialized in memory. 
We therefore design a fused GDN attention kernel that exploits cross-beam reuse of $S_{\mathrm{ctx}}$ and keeps projection replay entirely on chip.

\begin{figure}[t]
  \centering
  \includegraphics[width=\columnwidth]{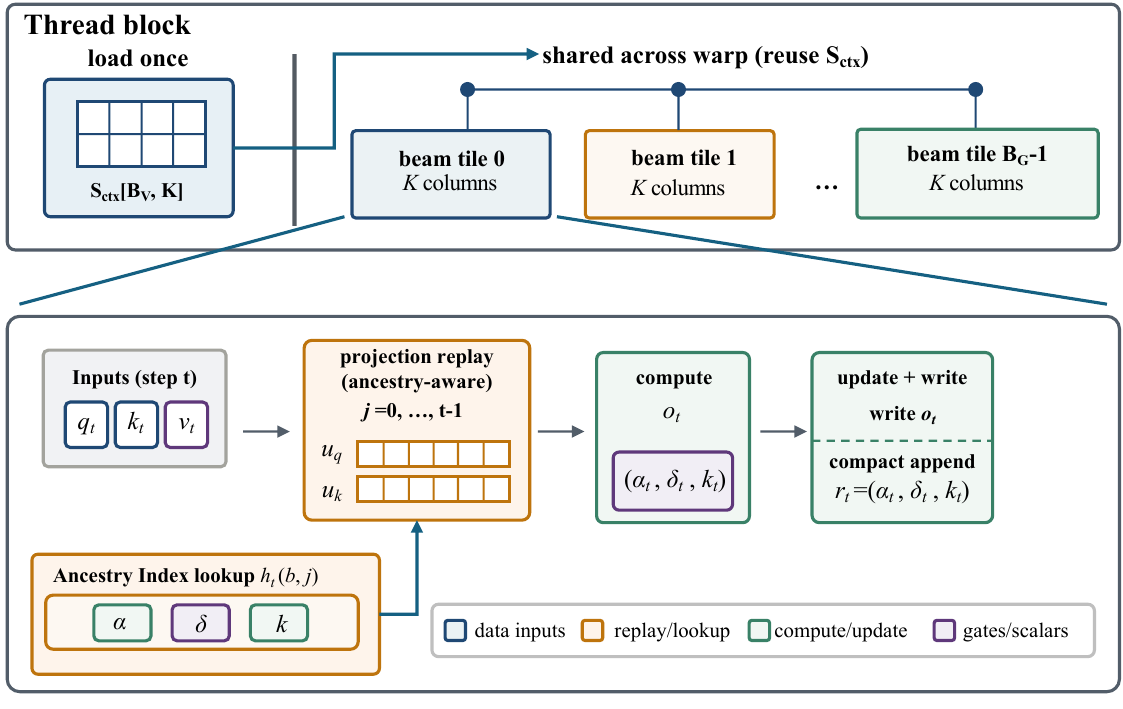}
  \vspace{-7mm}
  \caption{Shared-Context Attention Kernel}
  \Description{A three-panel single-column diagram highlights one program tile
  in the launch domain, expands its shared-base replay and output stores without
  full-state materialization, and shows the requested beam-tile pipeline.}
  \label{fig:temporal-kernel}
  \vspace{-7mm}
\end{figure}

To exploit cross-beam reuse, \sys assigns each GPU thread block a $B_V$-row tile in the value dimension together with a group of beams. 
As shown in Figure~\ref{fig:temporal-kernel}, the block loads the corresponding $S_{\mathrm{ctx}}[B_V,d_k]$ tile once into on-chip storage and reuses it across the beam tiles processed by the block. 
This amortizes the cost of reading $S_{\mathrm{ctx}}$ from HBM across multiple beams instead of incurring the same state read independently for every beam.

Within each thread block, the kernel fuses shared-context projection, ancestry-aware replay, and the current GDN update into one execution flow. For the current beam tile, the block first loads the current query, key, value, and gate inputs, and initializes two projection accumulators, $u_q = S_{\mathrm{ctx}} q_t$ and $u_k = S_{\mathrm{ctx}} k_t$, using the context state tile $S_{\mathrm{ctx}}[B_V,d_k]$.
It then replays the ancestral GTRs selected by the ancestry index in the following steps.
First, at replay depth $j$, the ancestry index $h_t(b,j)$ maps the logical ancestry position of beam $b$ to the corresponding physical GTR slot, from which the kernel gathers $r_j=(\alpha_j,\delta_j,k_j)$.
Second, the replay is applied directly to the two projection accumulators, updating them without ever reconstructing the intermediate recurrent state matrices $S_j$.
Third, the accumulators hold the required projections $S_{t-1}q_t$ and $S_{t-1}k_t$, which are then used immediately to compute the current transition record $r_t$ and output $o_t$.
Since the above steps are fused into a single kernel and no intermediate state matrix is reconstructed, the intermediate values remain on chip without being materialized in HBM.

Finally, the kernel appends only the compact GTR $r_t=(\alpha_t,\delta_t,k_t)$ for each newly generated beam, rather than writing back a full recurrent state $S_t\in\mathbb{R}^{d_v\times d_k}$. This eliminates the large per-beam state-writeback cost and preserves the compact state representation throughout decoding.

\section{Evaluation}
\label{sec:evaluation}

Our evaluation answers three questions:
(Q1) How much does \sys improve the end-to-end full-model decode performance?
(Q2) Does the GR GDN attention kernel maintain its latency and state-capacity advantages across workloads and model configurations? 
(Q3) How efficient is \sys's replay execution, and how much does it reduce memory traffic and consumption?

\subsection{Methodology}

We first describe the workload notation and experimental configurations used in the evaluation. 
Let $R$, $W$, and $O$ denote the request count, fixed logical beam width, and total number of output tokens, respectively. Prefill selects the first output token, leaving $T=O-1$ decode calls.
The full-model experiment uses the 0.8B Qwen3.5 model, with 6 full-attention layers and 18 GDN layers. We set $R$ to 4, $W$ to 256, the prompt length to 1K/5K, and $O$ to 3--7.
The operator suite covers 405 shapes over
$R=4/8/16$, $W=128/256/512$, $T=2$--$6$, and nine
recurrent geometries with $H_K=4$--32, $H_V=4$--64, $K=128$--256, and
$V=128$--512. On identical inputs and reorder-and-duplicate lineages, we pair
\sys with the unmodified vLLM/SGLang FullState and ReplaySSM kernels,
FlashInfer, and TensorRT-LLM FullState on their natively supported
shapes~\cite{kwon2023vllm,zheng2023sglang,ye2025flashinfer,trtllm}.

\vspace{-2mm}
\subsection{Full-Model Decode-Forward Performance}
\label{sec:eval-framework}

\begin{figure}[tbp]
  \centering
  \includegraphics[width=\columnwidth]{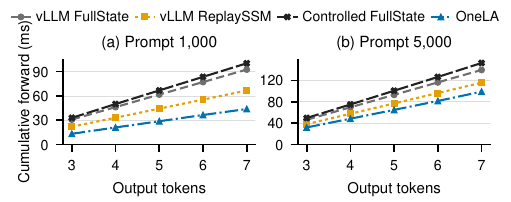}
  \vspace{-8mm}
  \caption{Cumulative full-model decode-forward GPU time.}
  \Description{Two line plots show cumulative decode model-forward GPU time
  versus output tokens for one-thousand- and five-thousand-token prompts.}
  \label{fig:framework}
  \vspace{-4mm}
\end{figure}

To show the end-to-end performance improvement when using \sys, Figure~\ref{fig:framework} reports the decode-forward GPU time of the complete model across the two prompt lengths.
We compare four execution paths. vLLM FullState and vLLM ReplaySSM serve as native framework references. Controlled FullState uses the same decode pipeline as OneLA while retaining conventional per-beam FullState execution, whereas OneLA is our full system.

\begin{figure}[tbp]
  \centering
  \includegraphics[width=\columnwidth]{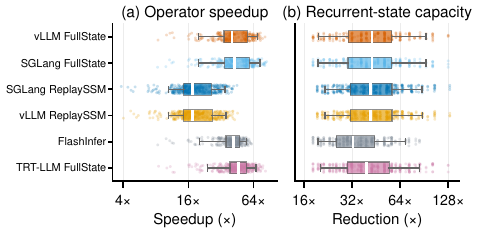}
  \vspace{-8mm}
  \caption{Operator speedup and persistent recurrent-state capacity reduction.}
  \Description{Two horizontal boxplots show OneLA's paired operator speedup
  and persistent recurrent-state capacity reduction over six baseline paths
  across all natively supported benchmark shapes.}
  \label{fig:generality}
  \vspace{-4mm}
\end{figure}

\begin{figure}[tbp]
  \centering
  \includegraphics[width=\columnwidth]{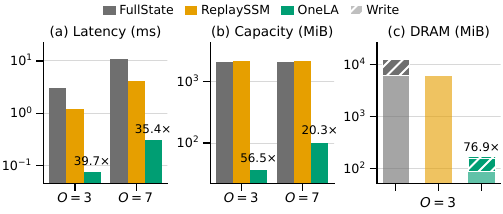}
  \vspace{-8mm}
  \caption{Recurrent-state efficiency breakdown.}
  \Description{Three panels show captured recurrent-state-path latency and
  persistent recurrent-state capacity at output three and seven, and
  NCU-measured physical DRAM reads and writes at output three.}
  \label{fig:temporal-efficiency}
  \vspace{-7mm}
\end{figure}

Controlled FullState remains within 6.9--9.3\% of native vLLM FullState, validating that our controlled implementation closely reproduces the framework baseline.
Replacing its recurrent state computation with \sys yields a 1.54--2.46$\times$ reduction in cumulative decode-forward time across all evaluated prompt and output lengths, and \sys consistently outperforms both vLLM execution paths.
These results show that \sys's recurrent-state optimization translates into substantial full-model decode-forward performance gains.

\subsection{GR GDN Attention Generality}
\label{sec:eval-generality}

We next isolate the GDN attention operator to evaluate whether \sys's latency and recurrent-state capacity advantages persist across diverse workloads and recurrent-state configurations.
For each paired workload, Figure~\ref{fig:generality} reports baseline/\sys ratios for captured latency and persistent recurrent-state capacity in GPU memory.
The evaluation covers all workloads natively supported by each baseline, where boxes indicate the median and interquartile range, and whiskers span the 5th--95th percentiles.

Figure~\ref{fig:generality}(a) shows that \sys consistently reduces GDN attention latency across the supported workloads.
Across all paired shapes, median speedups are 40.5$\times$ and 42.5$\times$ over vLLM and SGLang FullState, 17.5$\times$ over both ReplaySSM paths, and 41.5$\times$ and 46.3$\times$ over FlashInfer and TensorRT-LLM.
At the largest Qwen geometry ($R=16,W=512,H_K=H_V=16,K=V=128$), \sys takes 0.212~ms at $O=3$ and 0.990~ms at $O=7$, yielding 16.0--56.5$\times$ speedups over the four baselines supporting this shape. 
Across all 405 \sys cases, the maximum absolute output error against FP32 FullState is $4.88\times10^{-4}$, confirming numerical accuracy.

Figure~\ref{fig:generality}(b) shows similarly consistent reductions in persistent recurrent-state capacity.
Across the paired workloads, \sys reduces the median required capacity by up to 42.5$\times$ relative to the baselines, demonstrating that GTRs substantially reduce persistent GPU-memory requirements across different workload and recurrent-state configurations.
Together, these results show that \sys's GDN attention latency and recurrent-state capacity advantages generalize across request counts, beam widths, decode lengths, and all nine evaluated recurrent-state geometries.

\subsection{Sources of Performance Improvement}
\label{sec:eval-temporal-resources}

At the Qwen geometry with $R=4$ and $W=512$, we isolate the complete recurrent-state path and break down its performance through three complementary measurements: recurrent-path latency, persistent-state capacity and logical writes, and physical DRAM traffic.
Together, these measurements connect \sys's compact state representation to the reductions in memory movement that drive its execution-time improvement.

Figure~\ref{fig:temporal-efficiency}(a) first shows the resulting recurrent-state-path latency, including each path's explicit beam-state maintenance.
At $O=3$ and $O=7$, \sys takes only 0.0755 and 0.308~ms, whereas vLLM ReplaySSM takes 1.20 and 4.06~ms and vLLM FullState 3.00 and 10.9~ms, corresponding to 39.7$\times$ and 35.4$\times$ reductions relative to FullState.

Further, Figure~\ref{fig:temporal-efficiency}(b) shows that the reductions in persistent state and state writeback contribute to the improvement.
\sys requires only 36.3 and 101~MiB of persistent recurrent-state capacity at $O=3$ and $O=7$, corresponding to 56.5$\times$ and 20.3$\times$ reductions relative to FullState, respectively.
This reduction follows from replacing per-beam full-state matrices with one request-level $S_{\mathrm{ctx}}$ together with compact branch-specific GTRs.
The same representation also substantially reduces state writes: \sys writes only 32.3 and 96.8~MiB at $O=3$ and $O=7$, respectively, achieving a 127$\times$ reduction relative to FullState at both endpoints by appending compact GTRs instead of writing back full states.

Finally, Figure~\ref{fig:temporal-efficiency}(c) shows that the reduced state storage and writeback translate into substantially lower physical DRAM traffic.
At $O=3$, \sys moves only 161~MiB of data, exhibiting 76.9$\times$ and 39.9$\times$ reductions relative to FullState and ReplaySSM, respectively.
This reduction comes from cross-beam reuse of $S_{\mathrm{ctx}}$, which avoids repeated state reads and full-state writeback.
Consequently, \sys reaches an arithmetic intensity~\cite{williams2009roofline} of 40.0~FLOP/byte, 45.6$\times$ and 20.6$\times$ higher than FullState and ReplaySSM, respectively, directly explaining the recurrent-state-path latency improvement.

\FloatBarrier

\section{Conclusion}
\label{sec:conclusion}

We present \sys, an efficient linear-attention decoding framework for large-beam generative recommendation.
By replacing per-beam recurrent states with a shared post-prefill state, compact transition records, and lightweight ancestry, \sys enables on-chip projection replay and cross-beam state reuse.
\sys achieves 1.54--2.46$\times$ end-to-end speedups and substantial operator-level gains, demonstrating the effectiveness of compact state representation for dynamic large-beam decoding.

\newpage
\bibliographystyle{ACM-Reference-Format}
\bibliography{references}


\begin{thebibliography}{24}


\ifx \showCODEN    \undefined \def \showCODEN     #1{\unskip}     \fi
\ifx \showISBNx    \undefined \def \showISBNx     #1{\unskip}     \fi
\ifx \showISBNxiii \undefined \def \showISBNxiii  #1{\unskip}     \fi
\ifx \showISSN     \undefined \def \showISSN      #1{\unskip}     \fi
\ifx \showLCCN     \undefined \def \showLCCN      #1{\unskip}     \fi
\ifx \shownote     \undefined \def \shownote      #1{#1}          \fi
\ifx \showarticletitle \undefined \def \showarticletitle #1{#1}   \fi
\ifx \showURL      \undefined \def \showURL       {\relax}        \fi
\providecommand\bibfield[2]{#2}
\providecommand\bibinfo[2]{#2}
\providecommand\natexlab[1]{#1}
\providecommand\showeprint[2][]{arXiv:#2}

\bibitem[Chen et~al\mbox{.}(2026)]%
        {chen2026onesearch}
\bibfield{author}{\bibinfo{person}{Ben Chen}, \bibinfo{person}{Xian Guo},
  \bibinfo{person}{Siyuan Wang}, \bibinfo{person}{Zihan Liang},
  \bibinfo{person}{Yufei Ma}, \bibinfo{person}{Yue Lv}, \bibinfo{person}{Chenyi
  Lei}, \bibinfo{person}{Yuqing Ding}, \bibinfo{person}{Wenwu Ou},
  \bibinfo{person}{Han Li}, {and} \bibinfo{person}{Kun Gai}.}
  \bibinfo{year}{2026}\natexlab{}.
\newblock \showarticletitle{{OneSearch}: A Preliminary Exploration of the
  Unified End-to-End Generative Framework for E-commerce Search}. In
  \bibinfo{booktitle}{\emph{Proceedings of the 43rd International Conference on
  Machine Learning}} \emph{(\bibinfo{series}{Proceedings of Machine Learning
  Research}, Vol.~\bibinfo{volume}{306})}. \bibinfo{publisher}{PMLR}.
\newblock
\urldef\tempurl%
\url{https://openreview.net/forum?id=JKGgHY9FKa}
\showURL{%
\tempurl}


\bibitem[Guan et~al\mbox{.}(2026)]%
        {douyin_long_sequence}
\bibfield{author}{\bibinfo{person}{Lin Guan}, \bibinfo{person}{Jia-Qi Yang},
  \bibinfo{person}{Zhishan Zhao}, \bibinfo{person}{Beichuan Zhang},
  \bibinfo{person}{Bo Sun}, \bibinfo{person}{Xuanyuan Luo},
  \bibinfo{person}{Jinan Ni}, \bibinfo{person}{Xiaowen Li},
  \bibinfo{person}{Yuhang Qi}, \bibinfo{person}{Zhifang Fan},
  \bibinfo{person}{Hangyu Wang}, \bibinfo{person}{Qiwei Chen},
  \bibinfo{person}{Yi Cheng}, \bibinfo{person}{Feng Zhang}, {and}
  \bibinfo{person}{Xiao Yang}.} \bibinfo{year}{2026}\natexlab{}.
\newblock \showarticletitle{Make It Long, Keep It Fast: End-to-End 10k-Sequence
  Modeling at Billion Scale on {Douyin}}. In
  \bibinfo{booktitle}{\emph{Proceedings of the ACM Web Conference 2026}}.
  \bibinfo{publisher}{ACM}, \bibinfo{address}{New York, NY, USA},
  \bibinfo{pages}{7989--7998}.
\newblock
\href{https://doi.org/10.1145/3774904.3792811}{doi:\nolinkurl{10.1145/3774904.3792811}}


\bibitem[He et~al\mbox{.}(2026)]%
        {he2026plum}
\bibfield{author}{\bibinfo{person}{Ruining He}, \bibinfo{person}{Lukasz Heldt},
  \bibinfo{person}{Lichan Hong}, \bibinfo{person}{Raghunandan Keshavan},
  \bibinfo{person}{Shifan Mao}, \bibinfo{person}{Nikhil Mehta},
  \bibinfo{person}{Zhengyang Su}, \bibinfo{person}{Alicia Tsai},
  \bibinfo{person}{Yueqi Wang}, \bibinfo{person}{Shao-Chuan Wang},
  {et~al\mbox{.}}} \bibinfo{year}{2026}\natexlab{}.
\newblock \showarticletitle{Plum: Adapting pre-trained language models for
  industrial-scale generative recommendations}. In
  \bibinfo{booktitle}{\emph{Proceedings of the ACM Web Conference 2026}}.
  \bibinfo{pages}{8093--8104}.
\newblock


\bibitem[Hou et~al\mbox{.}(2025)]%
        {hou2025actionpiece}
\bibfield{author}{\bibinfo{person}{Yupeng Hou}, \bibinfo{person}{Jianmo Ni},
  \bibinfo{person}{Zhankui He}, \bibinfo{person}{Noveen Sachdeva},
  \bibinfo{person}{Wang-Cheng Kang}, \bibinfo{person}{Ed~H. Chi},
  \bibinfo{person}{Julian McAuley}, {and} \bibinfo{person}{Derek~Zhiyuan
  Cheng}.} \bibinfo{year}{2025}\natexlab{}.
\newblock \showarticletitle{{ActionPiece}: Contextually Tokenizing Action
  Sequences for Generative Recommendation}. In
  \bibinfo{booktitle}{\emph{Proceedings of the 42nd International Conference on
  Machine Learning}} \emph{(\bibinfo{series}{Proceedings of Machine Learning
  Research}, Vol.~\bibinfo{volume}{267})}. \bibinfo{publisher}{PMLR},
  \bibinfo{pages}{24004--24024}.
\newblock
\urldef\tempurl%
\url{https://proceedings.mlr.press/v267/hou25f.html}
\showURL{%
\tempurl}


\bibitem[Jiang et~al\mbox{.}(2026)]%
        {10.1145/3770855.3818473}
\bibfield{author}{\bibinfo{person}{Jie Jiang}, \bibinfo{person}{Hongbo Tang},
  \bibinfo{person}{Wenjie Wu}, \bibinfo{person}{Yangru Huang},
  \bibinfo{person}{ZhenMao Li}, \bibinfo{person}{Qian Li},
  \bibinfo{person}{Changping Wang}, \bibinfo{person}{Jun Zhang}, {and}
  \bibinfo{person}{Huan Yu}.} \bibinfo{year}{2026}\natexlab{}.
\newblock \showarticletitle{S-GRec: Personalized Semantic-Aware Generative
  Recommendation with Asymmetric Advantage}. In
  \bibinfo{booktitle}{\emph{Proceedings of the 32nd ACM SIGKDD Conference on
  Knowledge Discovery and Data Mining V.2}} (Republic of Korea)
  \emph{(\bibinfo{series}{KDD '26})}. \bibinfo{publisher}{Association for
  Computing Machinery}, \bibinfo{address}{New York, NY, USA},
  \bibinfo{pages}{7465–7476}.
\newblock
\showISBNx{9798400722592}
\href{https://doi.org/10.1145/3770855.3818473}{doi:\nolinkurl{10.1145/3770855.3818473}}


\bibitem[Katharopoulos et~al\mbox{.}(2020)]%
        {katharopoulos2020transformers}
\bibfield{author}{\bibinfo{person}{Angelos Katharopoulos},
  \bibinfo{person}{Apoorv Vyas}, \bibinfo{person}{Nikolaos Pappas}, {and}
  \bibinfo{person}{Fran{\c{c}}ois Fleuret}.} \bibinfo{year}{2020}\natexlab{}.
\newblock \showarticletitle{Transformers are {RNN}s: Fast Autoregressive
  Transformers with Linear Attention}. In \bibinfo{booktitle}{\emph{Proceedings
  of the 37th International Conference on Machine Learning}}
  \emph{(\bibinfo{series}{Proceedings of Machine Learning Research},
  Vol.~\bibinfo{volume}{119})}. \bibinfo{publisher}{PMLR},
  \bibinfo{pages}{5156--5165}.
\newblock
\urldef\tempurl%
\url{https://proceedings.mlr.press/v119/katharopoulos20a.html}
\showURL{%
\tempurl}


\bibitem[Kwon et~al\mbox{.}(2023)]%
        {kwon2023vllm}
\bibfield{author}{\bibinfo{person}{Woosuk Kwon}, \bibinfo{person}{Zhuohan Li},
  \bibinfo{person}{Siyuan Zhuang}, \bibinfo{person}{Ying Sheng},
  \bibinfo{person}{Lianmin Zheng}, \bibinfo{person}{Cody~Hao Yu},
  \bibinfo{person}{Joseph~E. Gonzalez}, \bibinfo{person}{Hao Zhang}, {and}
  \bibinfo{person}{Ion Stoica}.} \bibinfo{year}{2023}\natexlab{}.
\newblock \showarticletitle{Efficient Memory Management for Large Language
  Model Serving with {PagedAttention}}. In
  \bibinfo{booktitle}{\emph{Proceedings of the 29th Symposium on Operating
  Systems Principles}}. \bibinfo{publisher}{ACM}, \bibinfo{address}{New York,
  NY, USA}, \bibinfo{pages}{611--626}.
\newblock
\href{https://doi.org/10.1145/3600006.3613165}{doi:\nolinkurl{10.1145/3600006.3613165}}


\bibitem[{NVIDIA}(2026)]%
        {trtllm}
\bibfield{author}{\bibinfo{person}{{NVIDIA}}.} \bibinfo{year}{2026}\natexlab{}.
\newblock \bibinfo{title}{{TensorRT-LLM}}.
\newblock
\urldef\tempurl%
\url{https://github.com/NVIDIA/TensorRT-LLM}
\showURL{%
\tempurl}
\newblock
\shownote{Evaluated tag v1.3.0rc23 (d41ab33)}.


\bibitem[{OneRec Team} et~al\mbox{.}(2026)]%
        {onerec2026onereason}
\bibfield{author}{\bibinfo{person}{{OneRec Team}} {et~al\mbox{.}}}
  \bibinfo{year}{2026}\natexlab{}.
\newblock \bibinfo{title}{{OneReason} Technical Report}.
\newblock
\href{https://doi.org/10.48550/ARXIV.2606.06260}{doi:\nolinkurl{10.48550/ARXIV.2606.06260}}


\bibitem[{Qwen Team}(2025)]%
        {qwen3next}
\bibfield{author}{\bibinfo{person}{{Qwen Team}}.}
  \bibinfo{year}{2025}\natexlab{}.
\newblock \bibinfo{title}{{Qwen3-Next}: Towards Ultimate Training and Inference
  Efficiency}.
\newblock \bibinfo{howpublished}{Qwen Blog}.
\newblock
\urldef\tempurl%
\url{https://qwen.ai/blog?id=qwen3-next}
\showURL{%
\tempurl}
\newblock
\shownote{Accessed August 2026}.


\bibitem[Rajput et~al\mbox{.}(2023)]%
        {rajput2023tiger}
\bibfield{author}{\bibinfo{person}{Shashank Rajput}, \bibinfo{person}{Nikhil
  Mehta}, \bibinfo{person}{Anima Singh}, \bibinfo{person}{Raghunandan~Hulikal
  Keshavan}, \bibinfo{person}{Trung Vu}, \bibinfo{person}{Lukasz Heldt},
  \bibinfo{person}{Lichan Hong}, \bibinfo{person}{Yi Tay},
  \bibinfo{person}{Vinh~Q. Tran}, \bibinfo{person}{Jonah Samost},
  \bibinfo{person}{Maciej Kula}, \bibinfo{person}{Ed~H. Chi}, {and}
  \bibinfo{person}{Maheswaran Sathiamoorthy}.} \bibinfo{year}{2023}\natexlab{}.
\newblock \showarticletitle{Recommender Systems with Generative Retrieval}. In
  \bibinfo{booktitle}{\emph{Advances in Neural Information Processing
  Systems}}, Vol.~\bibinfo{volume}{36}. \bibinfo{publisher}{Curran Associates,
  Inc.}
\newblock
\href{https://doi.org/10.52202/075280-0452}{doi:\nolinkurl{10.52202/075280-0452}}


\bibitem[Sun et~al\mbox{.}(2025)]%
        {sun2025xgr}
\bibfield{author}{\bibinfo{person}{Qingxiao Sun}, \bibinfo{person}{Tongxuan
  Liu}, \bibinfo{person}{Shen Zhang}, \bibinfo{person}{Siyu Wu},
  \bibinfo{person}{Peijun Yang}, \bibinfo{person}{Haotian Liang},
  \bibinfo{person}{Menxin Li}, \bibinfo{person}{Xiaolong Ma},
  \bibinfo{person}{Zhiwei Liang}, \bibinfo{person}{Ziyi Ren},
  \bibinfo{person}{Minchao Zhang}, \bibinfo{person}{Yifan Wang},
  \bibinfo{person}{Xinyu Liu}, \bibinfo{person}{Ke Zhang},
  \bibinfo{person}{Hailong Yang}, {and} \bibinfo{person}{Depei Qian}.}
  \bibinfo{year}{2025}\natexlab{}.
\newblock \bibinfo{title}{{xGR}: Efficient Generative Recommendation Serving at
  Scale}.
\newblock
\href{https://doi.org/10.48550/ARXIV.2512.11529}{doi:\nolinkurl{10.48550/ARXIV.2512.11529}}


\bibitem[Vaswani et~al\mbox{.}(2017)]%
        {vaswani2017attention}
\bibfield{author}{\bibinfo{person}{Ashish Vaswani}, \bibinfo{person}{Noam
  Shazeer}, \bibinfo{person}{Niki Parmar}, \bibinfo{person}{Jakob Uszkoreit},
  \bibinfo{person}{Llion Jones}, \bibinfo{person}{Aidan~N. Gomez},
  \bibinfo{person}{Lukasz Kaiser}, {and} \bibinfo{person}{Illia Polosukhin}.}
  \bibinfo{year}{2017}\natexlab{}.
\newblock \showarticletitle{Attention Is All You Need}. In
  \bibinfo{booktitle}{\emph{Advances in Neural Information Processing
  Systems}}, Vol.~\bibinfo{volume}{30}. \bibinfo{publisher}{Curran Associates,
  Inc.}, \bibinfo{pages}{5998--6008}.
\newblock
\urldef\tempurl%
\url{https://papers.nips.cc/paper_files/paper/2017/hash/3f5ee243547dee91fbd053c1c4a845aa-Abstract.html}
\showURL{%
\tempurl}


\bibitem[Wang et~al\mbox{.}(2026)]%
        {adaptive_semantic_quantization}
\bibfield{author}{\bibinfo{person}{Huimu Wang}, \bibinfo{person}{Xingzhi Yao},
  \bibinfo{person}{Yiming Qiu}, \bibinfo{person}{Qinghong Zhang},
  \bibinfo{person}{Haotian Wang}, \bibinfo{person}{Yufan Cui},
  \bibinfo{person}{Songlin Wang}, \bibinfo{person}{Sulong Xu}, {and}
  \bibinfo{person}{Mingming Li}.} \bibinfo{year}{2026}\natexlab{}.
\newblock \bibinfo{title}{Towards Efficient and Generalizable Retrieval:
  Adaptive Semantic Quantization and Residual Knowledge Transfer}.
\newblock
\href{https://doi.org/10.48550/ARXIV.2602.23978}{doi:\nolinkurl{10.48550/ARXIV.2602.23978}}


\bibitem[Wang et~al\mbox{.}(2024a)]%
        {wang2024letter}
\bibfield{author}{\bibinfo{person}{Wenjie Wang}, \bibinfo{person}{Honghui Bao},
  \bibinfo{person}{Xinyu Lin}, \bibinfo{person}{Jizhi Zhang},
  \bibinfo{person}{Yongqi Li}, \bibinfo{person}{Fuli Feng},
  \bibinfo{person}{See-Kiong Ng}, {and} \bibinfo{person}{Tat-Seng Chua}.}
  \bibinfo{year}{2024}\natexlab{a}.
\newblock \showarticletitle{Learnable Item Tokenization for Generative
  Recommendation}. In \bibinfo{booktitle}{\emph{Proceedings of the 33rd ACM
  International Conference on Information and Knowledge Management}}.
  \bibinfo{publisher}{ACM}, \bibinfo{address}{New York, NY, USA},
  \bibinfo{pages}{2400--2409}.
\newblock
\href{https://doi.org/10.1145/3627673.3679569}{doi:\nolinkurl{10.1145/3627673.3679569}}


\bibitem[Wang et~al\mbox{.}(2024b)]%
        {wang2024eager}
\bibfield{author}{\bibinfo{person}{Ye Wang}, \bibinfo{person}{Jiahao Xun},
  \bibinfo{person}{Minjie Hong}, \bibinfo{person}{Jieming Zhu},
  \bibinfo{person}{Tao Jin}, \bibinfo{person}{Wang Lin},
  \bibinfo{person}{Haoyuan Li}, \bibinfo{person}{Linjun Li},
  \bibinfo{person}{Yan Xia}, \bibinfo{person}{Zhou Zhao}, {and}
  \bibinfo{person}{Zhenhua Dong}.} \bibinfo{year}{2024}\natexlab{b}.
\newblock \showarticletitle{{EAGER}: Two-Stream Generative Recommender with
  Behavior-Semantic Collaboration}. In \bibinfo{booktitle}{\emph{Proceedings of
  the 30th ACM SIGKDD Conference on Knowledge Discovery and Data Mining}}.
  \bibinfo{publisher}{ACM}, \bibinfo{address}{New York, NY, USA},
  \bibinfo{pages}{3245--3254}.
\newblock
\href{https://doi.org/10.1145/3637528.3671775}{doi:\nolinkurl{10.1145/3637528.3671775}}


\bibitem[Williams et~al\mbox{.}(2009)]%
        {williams2009roofline}
\bibfield{author}{\bibinfo{person}{Samuel Williams}, \bibinfo{person}{Andrew
  Waterman}, {and} \bibinfo{person}{David Patterson}.}
  \bibinfo{year}{2009}\natexlab{}.
\newblock \showarticletitle{Roofline: An Insightful Visual Performance Model
  for Multicore Architectures}.
\newblock \bibinfo{journal}{\emph{Commun. ACM}} \bibinfo{volume}{52},
  \bibinfo{number}{4} (\bibinfo{year}{2009}), \bibinfo{pages}{65--76}.
\newblock
\href{https://doi.org/10.1145/1498765.1498785}{doi:\nolinkurl{10.1145/1498765.1498785}}


\bibitem[Yang et~al\mbox{.}(2025)]%
        {yang2025gated}
\bibfield{author}{\bibinfo{person}{Songlin Yang}, \bibinfo{person}{Jan Kautz},
  {and} \bibinfo{person}{Ali Hatamizadeh}.} \bibinfo{year}{2025}\natexlab{}.
\newblock \showarticletitle{Gated Delta Networks: Improving {Mamba2} with Delta
  Rule}. In \bibinfo{booktitle}{\emph{The Thirteenth International Conference
  on Learning Representations}}.
\newblock
\urldef\tempurl%
\url{https://openreview.net/forum?id=r8H7xhYPwz}
\showURL{%
\tempurl}


\bibitem[Ye et~al\mbox{.}(2025)]%
        {ye2025flashinfer}
\bibfield{author}{\bibinfo{person}{Zihao Ye}, \bibinfo{person}{Lequn Chen},
  \bibinfo{person}{Ruihang Lai}, \bibinfo{person}{Wuwei Lin},
  \bibinfo{person}{Yineng Zhang}, \bibinfo{person}{Stephanie Wang},
  \bibinfo{person}{Tianqi Chen}, \bibinfo{person}{Baris Kasikci},
  \bibinfo{person}{Vinod Grover}, \bibinfo{person}{Arvind Krishnamurthy}, {and}
  \bibinfo{person}{Luis Ceze}.} \bibinfo{year}{2025}\natexlab{}.
\newblock \bibinfo{title}{FlashInfer: Efficient and Customizable Attention
  Engine for {LLM} Inference Serving}.
\newblock
\href{https://doi.org/10.48550/ARXIV.2501.01005}{doi:\nolinkurl{10.48550/ARXIV.2501.01005}}


\bibitem[Zhai et~al\mbox{.}(2024)]%
        {zhai2024hstu}
\bibfield{author}{\bibinfo{person}{Jiaqi Zhai}, \bibinfo{person}{Lucy Liao},
  \bibinfo{person}{Xing Liu}, \bibinfo{person}{Yueming Wang},
  \bibinfo{person}{Rui Li}, \bibinfo{person}{Xuan Cao}, \bibinfo{person}{Leon
  Gao}, \bibinfo{person}{Zhaojie Gong}, \bibinfo{person}{Fangda Gu},
  \bibinfo{person}{Jiayuan He}, \bibinfo{person}{Yinghai Lu}, {and}
  \bibinfo{person}{Yu Shi}.} \bibinfo{year}{2024}\natexlab{}.
\newblock \showarticletitle{Actions Speak Louder than Words: Trillion-Parameter
  Sequential Transducers for Generative Recommendations}. In
  \bibinfo{booktitle}{\emph{Proceedings of the 41st International Conference on
  Machine Learning}} \emph{(\bibinfo{series}{Proceedings of Machine Learning
  Research}, Vol.~\bibinfo{volume}{235})}. \bibinfo{publisher}{PMLR},
  \bibinfo{pages}{58484--58509}.
\newblock
\urldef\tempurl%
\url{https://proceedings.mlr.press/v235/zhai24a.html}
\showURL{%
\tempurl}


\bibitem[Zhao et~al\mbox{.}(2026)]%
        {holorec}
\bibfield{author}{\bibinfo{person}{Shuqi Zhao}, \bibinfo{person}{Jingsong Su},
  \bibinfo{person}{Xiang Liu}, \bibinfo{person}{Xingzhi Yao},
  \bibinfo{person}{Yiming Qiu}, \bibinfo{person}{Huimu Wang},
  \bibinfo{person}{Liang Lin}, \bibinfo{person}{Pengbo Mo},
  \bibinfo{person}{Mingming Li}, \bibinfo{person}{Jiao Dai},
  \bibinfo{person}{Jizhong Han}, {and} \bibinfo{person}{Songlin Hu}.}
  \bibinfo{year}{2026}\natexlab{}.
\newblock \bibinfo{title}{{HoloRec}: Holistic Encoding and Interleaved
  Reasoning for Generative Recommendation}.
\newblock
\href{https://doi.org/10.48550/ARXIV.2606.15331}{doi:\nolinkurl{10.48550/ARXIV.2606.15331}}


\bibitem[Zheng et~al\mbox{.}(2024)]%
        {zheng2024lcrec}
\bibfield{author}{\bibinfo{person}{Bowen Zheng}, \bibinfo{person}{Yupeng Hou},
  \bibinfo{person}{Hongyu Lu}, \bibinfo{person}{Yu Chen},
  \bibinfo{person}{Wayne~Xin Zhao}, \bibinfo{person}{Ming Chen}, {and}
  \bibinfo{person}{Ji-Rong Wen}.} \bibinfo{year}{2024}\natexlab{}.
\newblock \showarticletitle{Adapting Large Language Models by Integrating
  Collaborative Semantics for Recommendation}. In
  \bibinfo{booktitle}{\emph{2024 IEEE 40th International Conference on Data
  Engineering}}. \bibinfo{publisher}{IEEE}, \bibinfo{pages}{1435--1448}.
\newblock
\href{https://doi.org/10.1109/ICDE60146.2024.00118}{doi:\nolinkurl{10.1109/ICDE60146.2024.00118}}


\bibitem[Zheng et~al\mbox{.}(2023)]%
        {zheng2023sglang}
\bibfield{author}{\bibinfo{person}{Lianmin Zheng}, \bibinfo{person}{Liangsheng
  Yin}, \bibinfo{person}{Zhiqiang Xie}, \bibinfo{person}{Chuyue Sun},
  \bibinfo{person}{Jeff Huang}, \bibinfo{person}{Cody~Hao Yu},
  \bibinfo{person}{Shiyi Cao}, \bibinfo{person}{Christos Kozyrakis},
  \bibinfo{person}{Ion Stoica}, \bibinfo{person}{Joseph~E. Gonzalez},
  \bibinfo{person}{Clark Barrett}, {and} \bibinfo{person}{Ying Sheng}.}
  \bibinfo{year}{2023}\natexlab{}.
\newblock \bibinfo{title}{{SGLang}: Efficient Execution of Structured Language
  Model Programs}.
\newblock
\href{https://doi.org/10.48550/ARXIV.2312.07104}{doi:\nolinkurl{10.48550/ARXIV.2312.07104}}


\bibitem[Zhou et~al\mbox{.}(2025)]%
        {zhou2025onerec}
\bibfield{author}{\bibinfo{person}{Guorui Zhou}, \bibinfo{person}{Jiaxin Deng},
  \bibinfo{person}{Jinghao Zhang}, \bibinfo{person}{Kuo Cai},
  \bibinfo{person}{Lejian Ren}, \bibinfo{person}{Qiang Luo},
  \bibinfo{person}{Qianqian Wang}, \bibinfo{person}{Qigen Hu},
  \bibinfo{person}{Rui Huang}, \bibinfo{person}{Shiyao Wang}, {et~al\mbox{.}}}
  \bibinfo{year}{2025}\natexlab{}.
\newblock \bibinfo{title}{{OneRec} Technical Report}.
\newblock
\href{https://doi.org/10.48550/ARXIV.2506.13695}{doi:\nolinkurl{10.48550/ARXIV.2506.13695}}


\end{thebibliography}

\end{document}